# Max-Entropy Feed-Forward Clustering Neural Network

Han Xiao , Xiaoyan Zhu

*Abstract*—The outputs of non-linear feed-forward neural network are positive, which could be treated as probability when they are normalized to one. If we take Entropy-Based Principle into consideration, the outputs for each sample could be represented as the distribution of this sample for different clusters. Entropy-Based Principle is the principle with which we could estimate the unknown distribution under some limited conditions. As this paper defines two processes in Feed-Forward Neural Network, our limited condition is the abstracted features of samples which are worked out in the abstraction process. And the final outputs are the probability distribution for different clusters in the clustering process. As Entropy-Based Principle is considered into the feed-forward neural network, a clustering method is born. We have conducted some experiments on six open UCI datasets, comparing with a few baselines and applied purity as the measurement . The results illustrate that our method outperforms all the other baselines that are most popular clustering methods.

## I. Introduction

CLUSTERING is always popular in modern technology of artificial intelligence. It's a large branch of machine learning algorithms. No matter in theoretical or practical aspects, clustering is always fruitful and useful. The applications based on high quality clustering methods are so many, and most of them are important in either industrial or individual usages. For example, in the area of business, clustering machines help to find the users' group [1], and in the area of biology, they can help to discover genes and species. In natural language processing, clustering could help to discover the group of Morphologically Related Chinese Words [2]. Even in the task of discovering better conference paper keywords, clustering could be very useful [3]. Above, clustering is very important, and a better clustering machine could not only improve the basic belief of theories, but also promote the intelligent products.

Meanwhile, neural networks are also a large branch of learning algorithms, and at present the deeper architectures are becoming popular and attractive for promoting the ability of neural network. The neural network, especially feed-forward neural network, could extract the features and characteristics automatically from the original input data. However, traditional feed-forward neural networks are supervised learning machine that can only complete the tasks such as regression or classification, which means these methods need a lot of labelled data to train the network, while unsupervised learning tasks could hardly be completed by feed-forward neural networks. Though self-organized mapping or Kohenon network is also unsupervised learning machine, they should belong to shadow models which can not be improved by the deeper architecture. For these reasons, in this paper, we would like to propose an unsupervised multilayer feed-forward neural network which could benefit from both the extraction of features and unlabelled data.

In the learning process of supervised learning, the target function is based on the costs or losses of labelled data. However, when we deal with the problem of unlabelled data, what we hold are only the data manifolds. For this cause, we must select other principle, with which we could estimate the distribution under the conditions of each layers' data abstraction.

So in this paper, we have treated the output of neural layer as two kinds of information form. The outputs of each layer are both abstracted data features and probability distributions. For the latter form, Entropy-Based Principle is considered. In detail, the outputs of non-linear feed-forward networks are positive, which are the abstracted form of sample features and also could be viewed as probability distribution of different components in data manifold. When the Entropy-Based Principle is introduced into the feed-forward neural network, we maximize the entropy of clustering layers' probability distribution and minimize the entropy of abstraction layers' distribution. As to the inference of this neural network, we make the sample into the component which corresponds to the minimum output neuron. Noted that in our model, this minimum output neuron also means the maximum probability neuron. By this way, we could propose an unsupervised multilayer feed-forward neural network.

Our experiments are conducted on six UCI open datasets, that are Glass, Banknote Authentication, White Wine Quality, Red Wine Quality, Image Segment and MAGIC Gamma Telescope. We select four most popular clustering algorithms as baselines namely famous K-Means, density-based method, hierarchical clustering method and expectation maximum clustering method. The results of our experiments could prove the effectiveness of our method. Our experiments illustrate that our clustering method works well and outperforms common clustering methods. In conclusion, the Entropy-Based unsupervised learning algorithm based on feed-forward neural network would be reasonable and behave well.

Xiao is with the State Key Lab. of Intelligent Technology and Systems, National Lab. for Information Science and Technology, Dept. of Computer Science and Technology, Tsinghua University, Beijing 100084, PR China (e-mail:xiaoh12@mails.tsinghua.edu.cn)

Zhu is with the State Key Lab. of Intelligent Technology and Systems, National Lab. for Information Science and Technology, Dept. of Computer Science and Technology, Tsinghua University, Beijing 100084, PR China (e-mail:zxy-dcs@tsinghua.edu.cn)

The main contribution of this paper includes:

1) This paper studies two kinds of properties for the output of each neuron in network layers, which are abstracted data features and cluster probability distribution. With this theoretical analysis, the Entropy-Based Principle is applied into the feed-forward neural network to make it as an unsupervised learning machine which needs only unlabelled data.
2) An optimization problem is formulated based on above motivations. Clustering algorithm is designed as a solution of this optimization problem.

## II. BACKGROUND AND RELATED WORK

Clustering methods have always been a very huge branch of modern machine learning or data mining methods and it's one of the main topic of unsupervised learning. [4] and [5] had surveyed the algorithms of clustering methods. There are four kinds of most popular clustering algorithms. The first kind of popular clustering method is Distance-Based Clustering methods, which are focusing on the distance or similarity between samples or centroids. The famous K-Means and X-Means Clustering methods are included in this sort of algorithms as a special application of Vector Quantization, besides the graph-based methods are also popular methods which leverage the graph distance as the metric. The Fuzzy C-Means are based on Fuzzy Set Theories as its metric. Neural network model such as Self-Organized Mapping(SOM), and ARTS network also belong to this kind of clustering algorithm. The second kind of popular clustering method is the Hierarchical Clustering Methods, and this kind of methods are focusing on the iterative process to split different components in data manifolds. Methods such as Single linkage, Complete linkage, Group average linkage, Median linkage, Centroid linkage, Ward's method, Integer Linear Programming Hierarchical Clustering [6], BIRCH, CURE and ROCK are also in this kind of algorithms. The third popular kind of method is Density-Based Methods. The methods are based on the local metric between samples, which can discover many components in different shapes, styles and forms. The famous DBSCAN method, ADBSCAN [7] and OPTICS are also in this sort of clustering algorithms. The final kind of methods are based on the probabilistic viewpoint, the famous EM clustering method, mixed Gaussian Distribution, Bayesian non-parametric multilevel clustering [8] and Evolutionary soft co-clustering [9] pertain to this sort. Besides, the clustering methods based on information theory also belong to the final sort [10]. There are also some other clustering methods, while the above mentioned is the most popular ones. For an overview of ability compared between our model and common popular models, we select these four kinds of methods as our baselines.

Feed-Forward neural networks are also a large branch of machine learning algorithms, they have extraordinary data abstraction ability especially in the deep architectures. However, it is mainly used in supervised learning, which requires labelled data. The deep architecture could provide data abstraction ability and it could extract the features automatically. Thus the clustering tasks meeting the multilayer feed-forward neural network would become better than common neural network clustering methods, such as SOM and also better than common popular clustering methods.

Recently such stated in [11], going to deep catches many eyes, since not only just adding the hidden layers could gain an improvement in performance, but deep neural networks can also automatically select features and amazingly complete the comprehension missions. [12] had applied deep network into natural languages, and many works such as [13] and [14] had applied deep network into image processing. Obviously, deep learning is one of the hottest topic in today's machine learning theories and methods. Before [15] and [16] proposed the fast unsupervised and supervised methods, multi-layer neural networks are hard to train, this kind of difficulty is analysed and solved in [17] and [18], for the same cause our method could also perform well in deep architecture. Besides our work is a kind of principle, which could both works for shadow or deep architecture of feed-forward neural networks. In a word, we can cluster the data with deep or shadow architecture of Feed-Forward Neural Network, and a deeper architecture would benefit more.

Feed-forward neural network gains a powerful feature extraction ability, but few works could make it a clustering algorithm before this paper. As we have analysed about this situation, we conclude for two reasons. Firstly, traditional feed-forward neural network is the learning machine for the labelled data, and there is no principle or training target to make it able to learn from the unlabelled data. Based on this motivation, we propose Entropy-based principle which involves Min-Entropy and Max-Entropy Principle to overcome this difficult. Secondly, SOM and ARTs lead the road of neural clustering computing, it seems that only SOM or recursive neural network could be able to the clustering problem. However, with Entropy-based Principle, feed-forward neural network could also achieve this ability. Besides it outperforms most common clustering models with its amazing feature extraction ability.

Max-Entropy Principle is always used in probability distribution estimation where the distribution is unknown but is limited by some constrained conditions. The Max-Entropy Principle could be leveraged into logistic regression, and also be applied into feed-forward neural networks, which is proposed in this paper. The outputs of non-linear feed-forward neural networks are positive, thus they can be treated as both the abstraction of original data and the probability distribution of corresponding components in data manifolds. The abstraction of data manifold is the constrained condition, under which the outputs of each layer are the unknown probability distributions to be estimated. With this motivation, Feed-Forward neural networks with Max-Entropy Principle are able to cluster unlabelled data. Due to its strong ability of the abstraction process, it could be better than popular clustering methods. Our methods are both varified within theortical apsects and practical aspects.

## III. TWO PROCESSES IN FEED-FORWARD NEURAL NETWORK

### A. A Brief Illustration

In this section, we propose a novel viewpoint to review Feed-Forward Neural Network, which explains the network with two kinds of processes that are Abstraction and Clustering Process.

As the process that Fig. 1 illustrates, the iterative operations of feed-forward neural information processing could be treated as abstraction processes, where the features of sample are transducted from original feature space to abstracted feature space. In the stage of abstraction, each neuron would play a role as a linear regression learners, the new coordinated system is constructed by these linear regression learners with non-linear output function as its coordinates , and these linear regression learners could be viewed as the basis of the coordinated system for a new feature space.

As the abstraction process in Fig. 1 shows, the hidden neurons correspond to those dashed lines in original space, and the distance of samples to these lines construct the coordinated system of new feature space for the next process. The abstraction process could make the essence of data more easy and simple to be revealed. In this way, we could discover clusters in abstracted data space where the essence of data could be revealed more easily. When the cluster information is discovered, we can estimate the probability distribution of samples for different clusters in clustering process.

In the non-linear feed-forward neural network, the outputs of each layer have two properties. The first is abstracted feature which is discussed in above and the second is probability distribution for some components in data manifolds. As we know, each neuron catches some characteristics in data manifolds, which we could explain as that each neuron catches some kind of data clusters in data manifolds. For a sample in a specific cluster, the output neurons give out the probability distribution of corresponding clusters, based on which we could select the most possible cluster that the sample should belong to. The key of clustering process which could make the probability distribution for clusters in data manifold reasonable, lies to the training process, or we say the training principle. When we estimate the unknown probability distribution for data manifold, the Max-Entropy Principle works well, and the clustering algorithm is designed based on this principle.

### B. Abstraction Process in Min-Entropy Viewpoint

In the abstraction process, each neuron behaves as a linear regression learner, which catches part of data characteristics. This seems a little like the density-based methods, the abstracting neuron works as the density detector for data manifold. The output of each neuron could be treated as abstracted features or probabilistic distributions. The output of a neuron is smaller when a sample is more near to its corresponding regression hyper-plane, which illustrates that the linear regression learner works better. From the viewpoint of data abstraction, when we would like to enhance the abstracted ability, we must optimize a target which could make more samples near the only one corresponding linear regression learner. This part of target is supposed to have the following form.

Fig. 1. Two Processes in Clustering Feed-Forward Neural Networks.

$$Min \quad J = -\sum_{i=1}^{N_a}(\frac{1-O_i}{\sum_{j=1}^{N_a} 1-O_j})log(\frac{1-O_i}{\sum_{j=1}^{N_a} 1-O_j}) \quad (1)$$

In above formula, the $N_a$ is the neuron number in abstraction layer, and $O_i$ is the $i$-th unit output of this layer or we say the distance from sample to $i$-th linear regression learner, which is expressed in below part of this paper. In mathematical aspect, the target encourages that minority outputs of neurons become smaller and major outputs of neurons become bigger, which means that data samples get closer to one of the linear regression learners and the abstracted ability of the abstracting neurons becomes better. Thus minimizing this target, we make the data density detectors or we say our linear regression learners catch more essential data characteristics.

Also, we could review this target that we minimize the entropy of the abstraction processes in probabilistic or system viewpoint. As we know, the entropy of system means the uncertain level of the system, and in this situation, we want the data density could be analysed clearly enough to belong to only one determined linear regression learner, which means that we have to prefer a minimum of system uncertain level. For this reason, we must apply Min-Entropy Principle to abstraction process to gain the abstraction ability.

### C. Clustering Process in Max-Entropy Viewpoint

When the data density manifold could be analysed by lower layers which play the role of abstraction process, the output layer could behave as a clustering process, where we must determine how a sample belongs to those components that are detected from the abstraction process. This is a probabilistic estimation problem, in which the $1-O_i$ means the degree of the sample nearing the component as the same as the degree of component-sample membership. For this sake, we apply the Max-Entropy Principle to this estimation problem, with a distribution such like $(1-O_1, 1-O_2, 1-O_3...1-O_n)$. We obtain the target as following:

$$Max \quad J = -\sum_{i=1}^{N_c}(\frac{1-O_i}{\sum_{j=1}^{N_c}1-O_j})log(\frac{1-O_i}{\sum_{j=1}^{N_c}1-O_j}) \quad (2)$$

In above formula, the $N_c$ is the neuron number in clustering layer, and $O_i$ is the $i$-th unit output of this layer.

*D. Difficulties in Clustering And Solution in This Paper*

In a common viewpoint, there is a main difficulty in clustering algorithms that the shape of cluster or the flexibility of clustering model. Inflexible model can only characterize simple shape of cluster. For distance based clustering models, they own a fixed distance expression, which leads to inflexibility of clustering model. Meanwhile, for probabilistic clustering models, the assumption of cluster shape or we say the flexibility of model means very important to these methods. Thus, many clusters with different and peculiar shapes could not be detected well.

In this paper, our model could overcome this difficulty in clustering by abstraction and clustering process provided by feed-forward neural network. As we know, non-linear feed-forward neural network could express many kinds of functions which could be treated as many kinds of similarity or distance metric. For this point, our method provides a very flexible clustering model or a very various cluster shape assumption. In the abstraction process of our model, the data is transducted from one kind of extracted feature expression to another kind of extracted feature expression, and in the last kind of feature space, the data manifold or the cluster shape is very regular and easy to be revealed by Max-Entropy Principle. This point benefits from the motivation and results of the deep neural networks.

As we argued in this subsection, our model could both provide enough flexible clustering ability and detect many different and peculiar shapes. So our model is not limited by this kind of inflexible factor and overcomes this difficulty to achieve a better result.

IV. ALGORITHMS FOR MAX-ENTROPY FEED-FORWARD CLUSTERING NEURAL NETWORK

In the whole viewpoint, the last layer that the output layer behaves as a clustering layer, and other layers that are the hidden layers could be treated as abstraction layers. From the previous section, we could work out the whole target as followed:

$$Max \quad J = -\sum_{i=1}^{N}(\frac{1-O_i}{\sum_{j=1}^{N}1-O_j})log(\frac{1-O_i}{\sum_{j=1}^{N}1-O_j}) +$$
$$\lambda\sum_{l=1}^{L}\sum_{i=1}^{L_l}(\frac{1-O_{i,l}}{\sum_{j=1}^{L_l}1-O_{j,l}})log(\frac{1-O_{i,l}}{\sum_{j=1}^{L_l}1-O_{j,l}})$$

In above formula, the $N$ means the unit number of output layer, and the $L_l$ means the unit number of $l$-th hidden layer. The $L$ means the number of layers. The $O_i$ means $i$-th neuron output in output layer, and $O_{i,l}$ means the $i$-th neuron output in $l$-th hidden layer.

In this formula, we have the output of each layers as below:

$$O_j = \sigma(<\vec{w}_j^{out}, \vec{O}_L>) \quad (3)$$
$$O_{j,l} = \sigma(<\vec{w}_j^l, \vec{O}_{l-1}>) \quad (4)$$

In above two formulas, the $\sigma$ means the non-linear function applied into the clustering neural network and the $\vec{O}_l$ means the output vector of $l$-th layer that is composed by $O_j^l$ and $\vec{O}$ means the output vector of output layer. Besides, the operator $<\vec{x},\vec{y}>$ means an inner product operator between $\vec{x}$ and $\vec{y}$. In the vector-matrix form, we have following formulation.

$$<\vec{x},\vec{y}> = \vec{x}^T\vec{y}$$

The gradient descent optimization method is applied to solve the problem, so we must work out the derivative of the target, for brief illustration, we introduce the $\delta$ function.

$$\delta_{i,L}^{out} = \frac{O_i + N - 1 - \sum_{j=1}^{N}O_j}{(N - \sum_{j=1}^{N}O_j)^2}$$
$$(1 - log(\frac{1-O_i}{\sum_{j=1}^{N}1-O_j}))\sigma'(<\vec{w}_i^{out}, \vec{O}>) \quad (5)$$

$$\delta_{i,l-1}^{out} = \sum_{j=1}^{L_l}\delta_{j,l}^{out}\vec{w}_s^l(j)\sigma'(<\vec{w}_i^{l-1}, \vec{O}_{l-1}>) \quad (6)$$

$$\delta_{i,l}^l = \frac{O_{i,l} + L_l - 1 - \sum_{j=1}^{L_l}O_{j,l}}{(L_l - \sum_{j=1}^{L_l}O_{j,l})^2}$$
$$(1 - log(\frac{1-O_{i,l}}{\sum_{j=1}^{L_l}1-O_{j,l}}))\sigma'(<\vec{w}_i^l, \vec{O}_l>) \quad (7)$$

$$\delta_{i,l-1}^{l'} = \sum_{j=1}^{L_{l'}}\delta_{j,l}^{l'}\vec{w}_s^l(j)\sigma'(<\vec{w}_i^{l-1}, \vec{O}_{l-1}>) \quad (8)$$

In above formulas, the $\vec{w}$ is the weight vector for corresponding neuron, and $\vec{O}_l$ is the output vector of the $l$-th layer, which is composed by all the outputs of neurons in the $l$-th layer.

With the expression of $\delta$ function, we could reduce the complexity of computation, and express the updating equation very briefly.

$$\frac{\partial J}{\vec{w}_i^{out}} = \delta_{i,L}^{out}\vec{y}^L \quad (9)$$

$$\frac{\partial J}{\vec{w}_i^l} = \delta_{i,l}^{out}\vec{y}^l - \lambda\sum_{s=l}^{L}\delta_{i,l}^s\vec{y}^l \quad (10)$$

In above formula, $\sigma$ is the non-linear function and $\sigma'$ is the derivative form of non-linear function. Hence the updating equation is obtained as following:

$$\vec{w}_i^{out} = \vec{w}_i^{out} + \alpha\frac{\partial J}{\vec{w}_i^{out}} \quad (11)$$
$$\vec{w}_i^l = \vec{w}_i^l + \alpha\frac{\partial J}{\vec{w}_i^l} \quad (12)$$

In above equation, the $\alpha$ is the learning rate, so the clustering algorithm is obtained.

Our training algorithm as algorithm 1 shows, holds a

**Algorithm 1** Training Algorithm For Feed-Forward Clustering Neural Networks
   set randrom values to weights
   **repeat**
      To infer the neural network using Algortihm 2
      **for all** $\vec{w}_i^{out} \in$ output layer **do**
$$\vec{w}_i^{out} = \vec{w}_i^{out} + \alpha \frac{\partial J}{\vec{w}_i^{out}}$$
      **end for**
      **for all** $\vec{w}_i^l \in$ hidden layer **do**
$$\vec{w}_i^l = \vec{w}_i^l + \alpha \frac{\partial J}{\vec{w}_i^l}$$
      **end for**
   **until** All Samples' Convergence

computation complexity as

$$O(|NeuronNumber|^2 |LayerNumber|^2)$$

The complexity is proportional to the number of neurons in each layer and also the number of layers. For the reason that the number of layer is fixed and small, our algorithm could run as fast as common back-propagation neural network.

As previous section stated, the re-designed inference method is achieved as to modify the normal feed-forward neural network inference algorithm, in which we obtain the cluster that a sample belongs to by selecting the neuron which output the maximum of probability notated as $1 - O_i$. That is to say that we select the minimum output neuron as the cluster of our sample.

This re-designed inference algorithm or we say clustering algorithm is almost the basic back-propagation neural network inference algorithm. Only for the last step, we choose the minimum distance to the clustered component as our clustering choice. This algorithm could also run as fast as common back-propagation neural network inference method.

**Algorithm 2** Clustering Algorithm As Inference in Feed-Forward Clustering Neural Networks
   **for all** each sample $\vec{x}_t$ **do**
      **for all** each hidden neuron $\in$ hidden layer **do**
$$y_j = \sigma(<\vec{w}_j^l, \vec{o}^l>)$$
      **end for**
      **for all** each output neuron $\in$ output layer **do**
$$t_i = \sigma(<\vec{w}_i^{out}, \vec{o}^L>)$$
      **end for**
   **end for**
   **return** Order Number of Minimum Output of Neurons in Output Layers.

TABLE I.
The Purity of Different Datasets For Different Method

| Dataset | K-Means | Density | Hier. | EM | Ours |
|---|---|---|---|---|---|
| Glass | 0.54 | 0.48 | 0.37 | 0.53 | **0.63** |
| BankNote | 0.57 | 0.57 | 0.55 | 0.56 | **0.82** |
| White Wine | 0.45 | 0.44 | 0.42 | 0.45 | **0.49** |
| Red Wine | 0.48 | 0.48 | 0.40 | 0.46 | **0.53** |
| Image Segment | 0.53 | 0.55 | 0.16 | 0.55 | **0.61** |
| MAGIC | 0.49 | 0.52 | 0.59 | 0.59 | **0.69** |

## V. COMPARISON AMONG POPULAR CLUSTERING MODELS

The connection between distance based methods with our model is relatively obvious. The training algorithm could be treated as a process to achieve a better distance function which is encoded in the weights of neural networks. Then the inference and clustering process use this learned distance function that our neural network, to cluster data. There would be a fact that the distance of our model is trained from the dataset itself, with flexible form and reasonable training principle. So our model would not be easily affected by different shapes of cluster or fixed distance expression.

The connection between density-based methods with our model is also relatively obvious. Our method could be treated as a multi-layer or hierarchical density-based methods. One layer density based method could extract some clustering information, and this paper apply many layers density based method while each layer could extract more clustering information than the previous one. This process is just the abstraction process of neural network. With this process, our model that could be treated as hierarchical density-based method would be more precious than common density based methods.

The connection between hierarchical methods with our model is not very obvious. However, as we know, hierarchical methods also need a kind of fixed metric for linkage weights, which in our model is changed to a flexible and expressive distance formulation that is obtained by the training algorithm. For this key point, our model would gain the flexible distance adaptive to the specific dataset. Thus our model would be more competent than common hierarchical methods.

The connection between probabilistic methods with our model is relatively obvious. Our model is just a probabilistic model or information-theoretical model. However, common probability models are based on the original data, the features of which are rough. But this paper joints probabilistic models with neural network. Thus, our model would abstract rough features to refined features which could be easily analysed.

## VI. EXPERIMENTS

We have conducted experiments for the effectiveness of Feed-Forward Clustering Neural Networks. Each group of experiments has achieved good results to prove our methods and our theories are effective.

## A. Experimental Settings

We have selected six open UCI datasets, which are often used in clustering tasks.

**Banknote Authentication.** The data were extracted from images that were taken from genuine and forged banknote-like specimens and Wavelet Transform tool were used to extract features from images. there are 1,372 items with 5 attributes for binary classes.

**Glass.** The study of classification of types of glass was motivated by criminological investigation. At the scene of the crime, the glass left can be used as evidence. We use this dataset to testify our clustering method. There are 214 instances with 10 attributes for 6 classes.

**Red Wine Quality And White Wine Quality.** The two datasets are related to red and white variants of the Portuguese "Vinho Verde" wine. There are 1,599 instances with 11 attributes in the first dataset for 4 classes. And there are 4,899 instances with 11 attributes in the second dataset for 5 classes.

**Image Segmentation.** The instances were drawn randomly from a database of 7 outdoor images. The images were hand-segmented to create a classification for every pixel. There are 2,310 instances with 18 attributes for 7 classes.

**MAGIC Gamma Telescope.** It is generated to simulate registration of high energy gamma particles in a ground-based atmospheric Cherenkov gamma telescope using the imaging technique. There are 19,020 items with 11 attributes for binary classes.

## B. Effectiveness of Our method

For evaluation of effectiveness about our clustering algorithms, we choose four baselines, all of which are most famous and popular methods. Baselines and our model are listed as followed.

1) K-Means Algorithm, which is the distance based clustering methods, and the number of clusters is fixed, we try many settings for this methods, and work out the almost best purity results, and this method is implemented by WEKA, notated as K-Means.
2) Density-Based Algorithm, which is the density estimation clustering method, it can detect many shapes of clusters, but the number of clusters is fixed and this method is implemented by WEKA, notated as Density-Based. We also try many settings for this methods, and work out the almost best purity results.
3) Hierarchical-Based Algorithm, which is the hierarchical clustering method, the number of clusters is fixed, and this method is also implemented by WEKA, notated as Hierarchical. Many settings are tried for this method to work out the almost best purity results.
4) EM Algorithm, which must be provided with the number of cluster, and this method is also implemented by WEKA, notated as EM. This method is based on probabilistic principle. Different settings and parameters are tried to work out the almost best purity.
5) Our Algorithm, which is two layer clustering feed-forward neural network. Cluster number of our model is fixed, so we try some different settings to work out the almost best purity of it. and it is implemented by ourselves, and this model is notated as Ours.

Fig. 2. The beam means the purities of Glass for different hidden node number which is showed in x-axis. And The line means the target function values. The left y-axis value means target value and the right y-axis means purity.

Fig. 3. The beam means the purities of White Wine Quality for different hidden node number which is showed in x-axis. And The line means the target function values. The left y-axis value means target value and the right y-axis means purity.

We apply a popular evaluation method for clustering with classification data to testify our clustering algorithms. The purity is showed in below formula, and the higher the score is got, the better the system is.

$$Purity = \frac{1}{N} \sum_k \max_j |w_k \cap c_j| \quad (13)$$

In above formula, the $N$ is the number of instances. the $w_k$ is the set of samples in $k$-th cluster and the $c_j$ is the set of samples in $j$-th real class.

Table 1 shows the purity of different methods for different datasets. It' clearly and obversely that our model is better than others. The effectiveness about our model is verified, and the detailed analysis is listed as followed:

1) The reasons leading to the experimental result is analysed in Section V. That is to say, our theoretical analysis is verified.
2) Our method outperforms other popular clustering methods. The effectiveness of our method is verified,

## C. Network Structure Studies

In this subsection, we study the effect of the neural network structure.

*1) Experimental Settings:* We choose the classical three UCI datasets, which are Glass, White Wine Quality and Red Wine Quality, with different hidden node number to study the structure effect. Noted that the results on other datasets are similar in the aspect of trend. We also apply the purity that is defined above as our measurement.

*2) Results And Discussion:* The results are shown in the Fig. 2, 3 and 4, where more x-axis value is, more dense the neural network is. From the trend of these figures, we could conclude and explain some key points as following:

1) The purity of clustering has the same trend with target values while hidden node changes. We could choose the hidden node number based on this point.
2) Less hidden nodes often lead to unfitting problem while more hidden nodes often lead to overfitting problem. We shall chose a suitable hidden node number to avoid unfitting problem or overfitting problem.

Fig. 4. The beam means the purities of Red Wine Quality for different hidden node number which is showed in x-axis. And The line means the target function values. The left y-axis value means target value and the right y-axis means purity.

Our clustering feed-forward neural network may be stuck in unfitting problem and overfitting problem. However, we could use the trend to improve the clustering effect. But in the whole viewpoint, our method would also be robust enough to cluster the uncertain data.

In conclusion, our model is more flexible and more abstracting able than those common popular clustering methods. Jointing the abstraction ability of feed-forward neural networks and the probability estimation of Max-Entropy Principle works well.

## VII. Conclusion

In this paper, we propose a method to cluster, jointing Entropy-Based Principle with Feed-Forward Neural Networks. We treat the Feed-Forward Clustering Neural Network as two processes. In the abstraction process, Min-Entropy Principle is applied for more abstracted features and in the clustering process, Max-Entropy Principle is applied for distribution estimation of clusters in data manifold. We compare four kinds of popular clustering methods with our model, and conclude that our model would perform better for flexible metric representation and various shape assumption. Experiments are conduct on six open UCI datasets and four baselines are selected. Results show our model outperforms other common clustering methods and suitable hidden node selection would avoid unfitting problem or overfitting problem to perform better.